%% file: FermiChallenge.tex
\pdfoutput=1

\documentclass[11pt]{article}

\usepackage{emnlp2021}

\usepackage{times}
\usepackage{latexsym}
\usepackage{hyperref}
\hypersetup{
    colorlinks=true,
    linkcolor=blue,
    filecolor=magenta,      
    urlcolor=blue,
    pdftitle={Overleaf Example},
    pdfpagemode=FullScreen,
    }
\usepackage[T1]{fontenc}

\usepackage[utf8]{inputenc}

\usepackage{microtype}

\usepackage{graphicx}
\usepackage{mysymbols}
\usepackage{comment}
\usepackage{tabularx}
\usepackage{amsmath} 
\usepackage{amssymb}
\usepackage{pifont}
\usepackage{enumitem}
\usepackage{cleveref}
\usepackage{caption}
\usepackage{subcaption}
\usepackage{booktabs}
\usepackage{float}

\newcommand{\dset}{\texttt{Dataset-name}}
\newcommand{\ak}[1]{\textcolor{black}{#1}}
\newcommand{\rc}[1]{\textcolor{black}{#1}}
\newcommand{\pc}[1]{\textcolor{black}{#1}}
\newcommand{\ac}[1]{\textcolor{orange}{[AC: #1]}}

\newcommand{\xmark}{\ding{55}}

\newcommand{\eat}[1]{}
\usepackage{colortbl}

\usepackage{listings}
\usepackage{xcolor}
\usepackage{xspace}
\usepackage{multirow}

\newcommand{\fpscore}{\mathit{fp\_score}}
\newcommand{\realfp}{\textsc{RealFP}\xspace}
\newcommand{\synthfp}{\textsc{SynthFP}\xspace}
\newcommand{\PAns}{\mathit{PAns}}

\definecolor{codegreen}{rgb}{0,0.6,0}
\definecolor{codegray}{rgb}{0.5,0.5,0.5}
\definecolor{codepurple}{rgb}{0.58,0,0.82}
\definecolor{backcolour}{rgb}{0.95,0.95,0.92}

\lstdefinestyle{mystyle}{
    backgroundcolor=\color{backcolour},   
    commentstyle=\color{codegreen},
    keywordstyle=\color{magenta},
    stringstyle=\color{codepurple},
    basicstyle=\ttfamily\footnotesize,
    breakatwhitespace=false,         
    breaklines=true,                 
    captionpos=b,                    
    keepspaces=true,                 
    showspaces=false,                
    showstringspaces=false,
    showtabs=false,                  
    tabsize=2
}

\title{How Much Coffee Was Consumed During EMNLP 2019? \\
  Fermi Problems: A New Reasoning Challenge for AI}

\author{
  Ashwin Kalyan\textsuperscript{1} \hspace{2ex}
  Abhinav Kumar\textsuperscript{2} \hspace{2ex}
  Arjun Chandrasekaran\textsuperscript{3}\\
  \textbf{Ashish Sabharwal\textsuperscript{1} \hspace{2ex}
  Peter Clark\textsuperscript{1}} \\
  \hspace{1ex}\\
  \textsuperscript{1}Allen Institute for AI \hspace{2ex}
  \textsuperscript{2}Georgia Tech \hspace{2ex}
  \textsuperscript{3}Max Planck Institute for Intelligent Systems\\
}

\begin{document}

\maketitle

\begin{abstract}
Many real-world problems require the combined application of multiple reasoning abilities---employing suitable abstractions, commonsense knowledge, and creative synthesis of problem-solving strategies. To help advance AI systems towards such capabilities, we propose a new reasoning challenge, namely Fermi Problems (FPs), which are questions whose answers can only be approximately estimated because their precise computation is either impractical or impossible. For example, ``How much would the sea level rise if all ice in the world melted?'' FPs are commonly used in quizzes and interviews to bring out and evaluate the creative reasoning abilities of humans. 
To do the same for AI systems, we present two datasets: 1) A collection of 1k real-world FPs sourced from quizzes and olympiads; and 2) a bank of 10k synthetic FPs of intermediate complexity to serve as a sandbox for the harder real-world challenge.
In addition to question-answer pairs, the datasets contain detailed solutions in the form of an executable program and supporting facts, helping in supervision and evaluation of intermediate steps.
We demonstrate that even extensively fine-tuned large-scale language models perform poorly on these datasets, on average making estimates that are off by two orders of magnitude.
Our contribution is thus the crystallization of several unsolved AI problems into a single, new challenge that we hope will spur further advances in building systems that can reason.
\end{abstract}

\input{sections/intro}
\input{sections/fermi}

\input{sections/rel_work}

\input{sections/dataset_tasks}

\input{sections/experiments}
\input{sections/conclusion}

\begin{small}
\bibliography{custom}
\bibliographystyle{acl_natbib}
\end{small}

\clearpage
\appendix
\input{sections/appendix}

\end{document}

%% file: sections/intro.tex
\section{Introduction}

\begin{figure*}[t!]
\begin{center}
\includegraphics[width=1.0\linewidth]{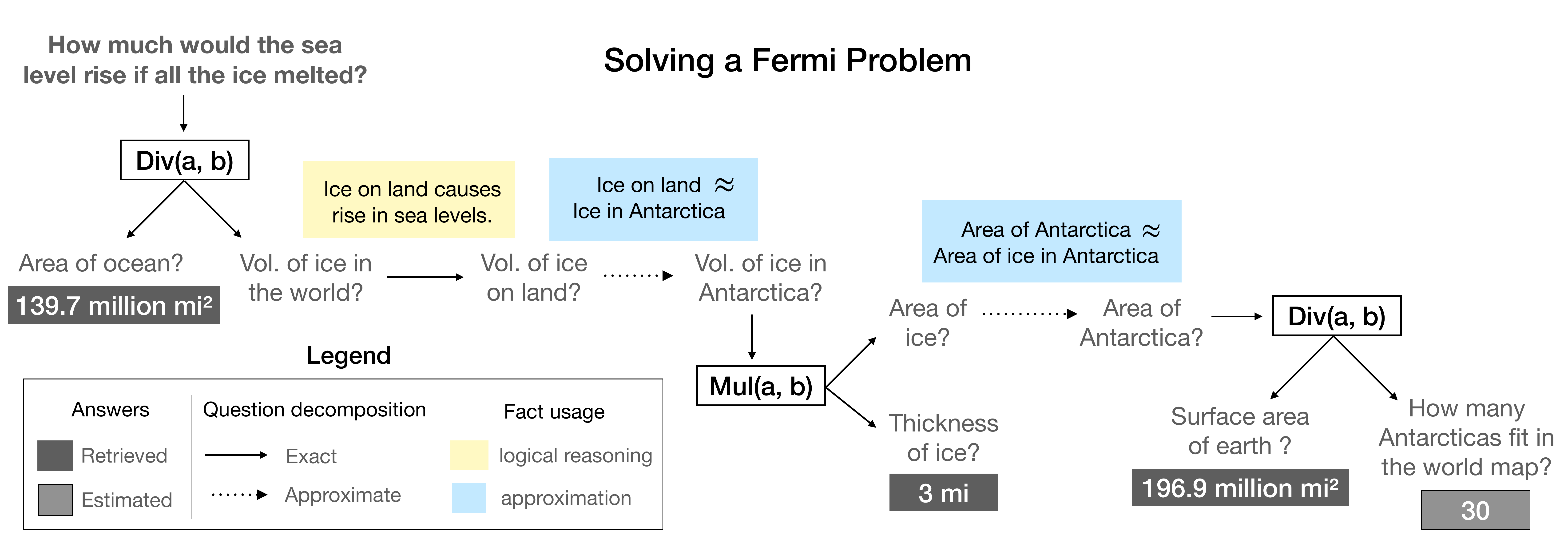}
\end{center}
\caption{
Humans solve FPs by employing sophisticated reasoning skills including abstraction, (ice on land $\approx$ ice on Antarctica),  problem decomposition (Volume of ice $=$ Area of Ice $\times$ thickness of ice) and commonsense reasoning (only ice on land causes rise in sea levels). 
    }
\vspace{-10pt}
\label{fig:teaser}
\end{figure*}



\emph{How long is the drive from Seattle to NYC?}
\emph{How big of an emergency fund do I need?}
We frequently encounter such questions in our daily lives. Likewise, scientists are often faced with questions such as,
\emph{How much would the ocean surface rise if the ice caps melted?}
Known as \emph{Fermi Problems}\footnote{after the celebrated physicist Enrico Fermi. See \url{https://en.wikipedia.org/wiki/Fermi_problem}.} (FPs), these questions are problems whose answers can only be estimated within \emph{reasonable} limits of error, as precisely measuring the required quantity is either impossible or impractical. 

Solving a FP requires considerable life experience, ability to think through long chains of reasoning, and mathematical intuition -- as illustrated by \cref{fig:teaser} for a FP about rising sea levels.
Answering an FP correctly requires multiple different facts, and the correct estimate can be arrived via various reasoning paths -- this open-ended nature further adds to their challenge. 
Unsurprisingly, these questions are often used to test candidates in science Olympiads and interviews. 
Due to the complexity of reasoning that is required to answer these questions, \emph{we propose solving FPs as a new task to drive progress in AI reasoning systems.}

A core skill required for solving FPs is that of \emph{estimation} -- \eg ``How much money do I need for a medical emergency?'', ``What is the thickness of ice sheets in Antarctica'', etc. 
This crucially requires \emph{abstracting} out details to simplify a complex real-world problem, similar to the (in)famous metaphor of the spherical cow,\footnote{\url{https://en.wikipedia.org/wiki/Spherical\_cow}}
For instance, when estimating the volume of Mt.~Everest\footnote{FP: How many dump trucks are needed to empty Mt.~Everest?}, abstracting that the mountain is a conical shaped object significantly simplifies the estimation problem at hand.
To the best of our knowledge, our proposed FP challenge is the first of its kind that requires reasoning of this nature.
Through this challenge, we hope to spur research towards building AI reasoning systems that are capable of performing such abstractions, a key reasoning skill that is 
natural to humans. 

The complex reasoning involved in solving FPs (see \cref{fig:teaser}) often means the question must be creatively decomposed into simpler ones. 
These simpler questions often themselves are open-ended Fermi problems. 
We thus hope our challenge will encourage advances in recursive reasoning models.

Further, FPs require the combined application of multiple reasoning strategies to solve the problem.
Unlike existing datasets and tasks geared towards specific reasoning skills (\eg commonsense reasoning or question decomposition), 
we hope our work drives progress in not just the ability of AI to employ suitable abstractions and estimations, but also in models that can combine various reasoning abilities to produce a coherent solution.

\paragraph{Contributions.} 
\begin{enumerate}[leftmargin=10pt, noitemsep]
    \item We introduce Fermi Problems (FPs), as a task to drive progress in AI reasoning systems -- specifically, testing for their ability to make reasonable abstractions, creatively decompose questions into solvable chunks and employ commonsense reasoning.
    
    \item We collect a set \realfp of 1k real-world FPs aggregated from numerous websites, quizzes and Olympiads.  Further, we provide a synthetic dataset \synthfp of 10k questions with the aim of serving as a bank of more accessible problems of intermediate complexity -- and hopefully, aid the development of AI models for the harder real-world setting. Both datasets are available at \url{https://allenai.org/data/fermi}.
    
    \item Based on the FP datasets, we propose three tasks of increasing hardness and establish baselines built around state-of-the-art language models.
    We find that FPs are well beyond the reach of such systems even after substantial fine-tuning -- on average, making predictions that are off by two orders of magnitude and only slightly better than predicting a constant value.
    Further, we provide an analysis of both the dataset and baselines to illustrate the hardness of the proposed tasks and motivate future advances.
\end{enumerate}

%% file: sections/fermi.tex
\section{Fermi Problems}
\label{sec:fermi}

The following properties of FPs and their solutions make them an ideal candidate for evaluating and advancing AI reasoning -- 
\\ \\
\noindent \emph{(1) Recursive Nature of Sub-Problems.} 
    As mentioned previously, problem decomposition is an important aspect of FPs. 
    An interesting property of FPs is that decomposed sub-problems are also FPs -- \eg ``How many dump trucks to empty Mt. Everest?'' requires answering the -- ``What is the volume of Mt. Everest?'' and ``What is the volume of a dump truck?'', which are in-turn, FPs. 
We employ this property of FPs to create a richer synthetic dataset (see \cref{sec:datasets-tasks} for more details).
    \\ \\
    \emph{(2) Creativity in FP solutions.} Problem decomposition for FPs is not only recursive but also requires considerable amount of creativity. 
    For the above FP about emptying Mt. Everest, an alternative decomposition is -- ``How many dump trucks to empty Mt. Rainier?'' and ``How many Mt. Rainiers fit in Mt. Everest?''. 
    Note that the decomposition still retains the recursive nature but now follows an alternate path.
    The exact decomposition is a function of the knowledge and life experiences of a person, and in the case of an AI, the information accessible to it -- either through information stored in its parameters, a retrieval mechanism or a knowledge base. As an accurate estimate is sought at the end, creativity in problem decomposition is closely intertwined with the problem of what can be estimated.
    In addition to practical scenarios (\eg ``How many port-a-potties are needed for a gathering of 1 million people?''), FPs often concern (a) unrelated objects (\eg Mars bars and Olympic pools), (b) unusual attributes of common objects (\eg volume of a Mars Bar as opposed to its calorific value) and (c) hypothetical scenarios (\eg ``Consider the earth and moon are at two ends of the school oval, how far is the sun?"). 
    Thus, FPs require going beyond biases encountered in the real world or in previous problems.\footnote{Perhaps, this is the reason that Fermi problems stump humans, especially when asked in situations like interviews.}
    Estimating the answer to such questions requires the ability to think creatively, a thorough understanding of the underlying process and the intent of the question.
\\ \\
\noindent \emph{(3) Need for Reasonable Abstractions.} 
    Despite taking a creative approach, one can be unsuccesful at solving FPs without the ability to make \emph{reasonable abstractions.}
    Returning to our running example of emptying Mt. Everest, a creative decomposition leads us to considering the volume of Mt. Everest \wrt to that of Mt. Rainier. 
    However, we still need to address the issue of computing the volume of Mt. Rainier -- here, assuming it to be a conical shaped object helps us in computing a reasonable estimate. 
    We humans employ various abstractions regularly in our daily lives -- \eg spatial abstraction (``Is the road wide enough to turn my car?''), temporal (``Do I have enough time to grab lunch before the next meeting?'') and causal (``Pressing the gas pedal, makes my car rush forward''). 
    We would require such a key skill to be well within the reach of AI systems and to this end, the proposed FP challenge is an ideal downstream task to evaluate this. 
\\ \\
\noindent \emph{(4) Commonsense Reasoning.} Arriving at the correct answer requires one to make reasonable abstractions at each step.
    This requires a sufficiently accurate working model of world and is broadly categorized as \emph{life experience}. 
For example -- the fact that a Mars Bar can be eaten in a few bites can help determine its volume. 
    Similarly, understanding that pizza shops usually cater to homes within a few mile radius helps in estimating the number of pizza delivery persons in Chicago. 
    Further, domain-specific reasoning might be required to solve some FPs -- for example FP illustrated in \cref{fig:teaser} requires an understanding of physics to infer that only land ice leads to increase in sea levels. 

%% file: sections/rel_work.tex
\eat{
\begin{table*}[t!]
\centering
{\small
\begin{tabular}{p{0.3\textwidth} p{0.45\textwidth} p{0.1\textwidth}}
\hline
\textbf{Dataset} & \textbf{Reasoning type} & \textbf{Requires knowledge retrieval?} \\
\hline
\texttt{DROP} \cite{dua-etal-2019-drop} & counting, addition, sorting, etc. &  \xmark \\
\texttt{HOTPOTQA} \cite{yang-etal-2018-hotpotqa} &  comparison, ..., etc. & \checkmark \\
\texttt{QAngaroo} \cite{10.1162/tacl_a_00021} & ? & \xmark \\
\texttt{StrategyQA} \cite{geva2021strategyqa} & comparison \pc{SQA has more than this!} & \checkmark \\
\texttt{ComplexWebQuestions} \cite{talmor-berant-2018-web} & superlatives, comparatives, union, intersection, addition & \checkmark \\
\texttt{MultiRCcorpus} \cite{khashabi-etal-2018-looking} & implied counts, sentiments, relationships & \xmark \\ 
\texttt{QASC} \cite{Khot_Clark_Guerquin_Jansen_Sabharwal_2020}  & common-sense reasoning & \checkmark \\ 
\texttt{OpenBookQA} \cite{Mihaylov2018CanAS} &  world-knowledge, common-sense reasoning & \checkmark \\ 
\hline 
\dset & world-knowledge, common-sense, mathematical reasoning & \checkmark \\ 
\hline 
\end{tabular}
}
\caption{Comparison of \dset~with existing datasets. \dset~uniquely contains `complex' questions that typically require mathematical reasoning in order to answer. \ac{TODO: Rethink column-headings, verify all entries, add missing datasets.}
}
\label{tab:dataset-comparison}
\end{table*}
}
  
\section{Related Work}



\paragraph{Mathematical Reasoning}
In the area of mathematical reasoning, several projects have probed the limits of transformers
to solve pure math problems \cite{saxton2019analysing, lample2019deep, hendrycks2021measuring}.
FPs differ from these problems in two important ways.
First, due to the heuristic nature of their solutions, FPs do not have a unique, precise answer with formal proof, in the way that
normal mathematical problems do. Second, FPs are stated in natural language (NL) rather
than a formal, mathematical notation. FPs are perhaps closer to algebra word
problems, where a NL question, e.g., ``How many cookies were left?'', 
is asked about a simple NL story \cite{Amini2019MathQATI,Ling2017ProgramIB,KoncelKedziorski2015ParsingAW}.
However, in algebra word problems, 
answers are again uniquely defined and provable. In addition, all required
information is provided in the story, while in FPs the solver must find/recall
required information.\footnote{
  We later also define a simpler FP task in which the required information is provided.}
Finally, in story problems, the space of possible solution equations is typically small and well-defined enough
that it can be exhaustively searched, while FPs can have arbitrarily complex solutions
(e.g., Figure~\ref{fig:teaser}).

\paragraph{Question Decomposition}
FPs require problem decomposition, in a way loosely similar to multihop inference.
However, for FPs, the appropriate decomposition is {\it not} explicit in the question
itself, unlike early multihop datasets such as HotpotQA \cite{Yang2018HotpotQAAD}
or WebQuestions \cite{Berant2013SemanticPO}. Later multihop datasets,
e.g., OBQA \cite{Mihaylov2018CanAS}, contained questions where the decomposition
was not explicit in the question (e.g., ``Does a suit of armor conduct electricity?'',
implicitly requiring a subquestion about materials), but typically into
just two (or at most three) steps. In contrast, FPs typically require multiple
levels of decomposition, significantly increasing complexity. This in turn
requires identifying a solution {\it strategy}, namely how to factor an unknown
quantity into a function of known (or recursively factorable) quantities.
The StrategyQA \cite{geva2021strategyqa} dataset illustrates this problem
but for a different task, namely true/false questions about whether something
is possible, and without recursive decomposition, a key feature of FPs.

\paragraph{Commonsense}
In addition to mathematical reasoning, FPs require significant commonsense
knowledge, both for estimating quantities and for decomposing problems.
For example, ``How many pizza delivery trucks are in Chicago?'' requires
significant commonsense about human behavior (How often do people
order pizza? How many deliveries can a truck make per day?) to even
begin to decompose the problem, let alone estimating basic quantities
(Population of Chicago?). While new resources of commonsense knowledge
are becoming available, e.g., \cite{Bosselut2019COMETCT,Sap2019ATOMICAA},
substantial development is still needed for the kind of world modeling that many
FPs require.

\paragraph{Numeric Estimation}
Large-scale language models trained on web-scale data have
been shown to contain common numerical facts -- \eg number of days in a year,
distance from earth to moon, number of hairs on a human head, etc.
We leverage one such model (T5 \cite{raffel2019exploring}) for our baselines.
More recently, researchers have shown that models can also perform
estimation to some degree \cite{zhang-etal-2020-language-embeddings},
and have proposed novel encoding strategies to improve number prediction
and estimation \cite{spithourakis2018numeracy,berg-kirkpatrick-spokoyny-2020-empirical}.
Such techniques would be valuable for improved solutions to FPs.

%% file: sections/dataset_tasks.tex
  \eat{
\begin{figure*}[t]
    \centering
    \begin{lstlisting}[frame=single]
    Q0: How many litres of water does the school use each week?
    
    Q1: What is the duration of the week in days?
    Q2: What is the total water consumption by students in the school?
    Q3: What is the individual water consumption of a student?
    Q4: What is the total number of students in the school?
    
    F1: There are 7 days in a week.
    F2: An average student's water consumption per day amounts to 18 litres
    F3: The total number of students in the school is 516
    
    A1: 7
    A2: 18 L
    A3: 516
    
    Q0 -> Mul (Q1, Q2)
    Q1 -> A1 | F1
    Q2 -> Mul (Q3, Q4)
    Q3 -> A2 | F2
    Q4 -> A3 | F3
    \end{lstlisting}
    \caption{\ak{Example data from \realfp dataset}}
    \label{fig:real_example}
\end{figure*}
}

\section{Datasets and Tasks}
\label{sec:datasets-tasks}

We present two datasets, \realfp and \synthfp, which are collections of real-world and synthetic Fermi problems, respectively.
We then define three FP challenge tasks, with varying difficulty levels.

\subsection{Dataset Elements}
\label{subsec:dataset-elements}

Each instance in our datasets consists of a Fermi question $Q$ and its answer $A$, standardized using the International System of Units, SI.\footnote{{\tiny \url{https://en.wikipedia.org/wiki/International\_System\_of\_Units}}}
Further, we add two extra elements to each question $Q$, supporting facts and explanations.

\paragraph{Supporting Facts $F$:}
Each question $Q$ is paired with $F$, a set of supporting facts, which are 
sentences describing quantities relevant to $Q$.
This enables two aspects of our Fermi challenge: (a) defining certain tasks where
the output must include $F$ as part of an explanation, to encourage justifiable
reasoning (see below); and (b) defining simpler FP tasks where $F$ (or a noisy version of it), is provided as part of the input
(as question ``context'') to help drive progress on the FP challenge under the familiar Reading Comprehension setting.

\begin{figure}[t]
\centerline{
 \fbox{%
   \parbox{\columnwidth}{
  {\it program} $\rightarrow$ {\it statement*} \\
  {\it statement} $\rightarrow$ {\it comp-expr} | {\it support-expr}  
   \vspace*{1mm} \\
  {\it comp-expr} $\rightarrow$ {\it qn-id} "{\tt ->}" \{{\it math-expr} | {\it value-expr}\} \\
  {\it math-expr} $\rightarrow$ {\it operator} "(" {\it qn-id}* ")" \\
  {\it operator} $\rightarrow$ "Add" | "Sub" | "Mul" | "Div" \\
  {\it value-expr} $\rightarrow$ {\it val-id} "because" {\it fact-id} 
  \vspace*{1mm} \\
  {\it support-expr} $\rightarrow$ {\it question-expr} | {\it fact-expr} | {\it val-expr} \\
  {\it question-expr} $\rightarrow$ {\it qn-id} ": " {\it question} \\
  {\it fact-expr}$^\dagger$ $\rightarrow$ {\it fact-id} ": " {\it sentence} \\
  {\it val-expr} $\rightarrow$ {\it val-id} ": " {\it number} [{\it units}]
}}}
  \caption{
    Grammar for FP explanation programs. $^\dagger$The proposed FP tasks (proposed in \cref{subsec:tasks}) separate out {\it fact-expr} from the program to either provide them as part of the input or expect them in the output.
    \label{grammar}
  }
  \vspace{-10pt}
\end{figure}
  
\begin{figure}[t]
\centerline{
 \fbox{%
   \parbox{1\columnwidth}{
     \small{
     \underline{Inputs:} \\
     {\bf Q:} "How many litres of water does a school use each week?" \\
     {\bf F:} "F1: There are 7 days in a week.\\
    \hspace*{4mm} F2: An average student's water consumption per day \\
    \hspace*{9mm} amounts to 18 litres.\\
    \hspace*{4mm} F3: The total number of students in a school is 516." \\
     \\
     \underline{Outputs:} \\
     {\bf A:} 65016 L\\
     {\bf E:} \\
    \hspace*{0.1mm} "Q0: How many litres of water does a school use each week?\\
    \hspace*{1mm} Q1: What is the duration of the week in days?\\
    \hspace*{1mm} Q2: What is the daily water consumption in the school?\\
    \hspace*{1mm} Q3: What is the daily water consumption of a student?\\
    \hspace*{1mm} Q4: What is the total number of students in the school?\\
    \\
    \hspace*{1mm} A1: 7 \\
    \hspace*{1mm} A2: 18 L \\
    \hspace*{1mm} A3: 516\\
    \\
    \hspace*{1mm} Q0 $\rightarrow$ Mul (Q1, Q2)\\
    \hspace*{1mm} Q1 $\rightarrow$ A1 because F1\\
    \hspace*{1mm} Q2 $\rightarrow$ Mul (Q3, Q4)\\
    \hspace*{1mm} Q3 $\rightarrow$ A2 because F2\\
    \hspace*{1mm} Q4 $\rightarrow$ A3 because F3 "
   }}}}
    \caption{\label{fig:real_example1} Example I/O for Task 2. The input is a Fermi question $Q$ and relevant facts $F$. The output is the answer $A$ and an explanation $P$ in the form of a program.
    }
    \vspace{-10pt}
\end{figure}

\paragraph{Explanations $P$:} 
In the case of FPs, the reasoning behind an answer is as important as the answer itself and therefore,
each question is paired with an explanation in the form of an \emph{executable program}
describing the facts, values, and mathematical computations needed to arrive at an answer -- see \cref{fig:real_example1} for an example.
The explanation programs that can be expressed are captured by a simple, recursive grammar shown in \cref{grammar}.

As seen from the grammar, an FP program is a sequence of statements, where each statement is either a computation expression or a support (or explanation) expression. A computation expression can be either a mathematical operator applied to one or more recursively spawned sub-questions (e.g., Q0 $\rightarrow$ Mul(Q1, Q2) in Figure~\ref{fig:real_example1}), or a value expression pointing to the identifier of a numerical value along with the identifier of a fact supporting that value (e.g., Q1 $\rightarrow$ A1 because F1). A support expression defines a sub-question, supporting fact, or numerical value, and associates it with a unique identifier for reference in the rest of the program (e.g., Q1: What is $\dots$, F1: There are $\dots$, and A1: 7 in the example in Figure~\ref{fig:real_example1}).

A program $P$ that respects this grammar can be ``executed'' or evaluated to obtain a numerical answer, using only the computation and value expressions contained in $P$. The sub-question and fact expressions included in $P$ act as provenance for the numerical computation captured by $P$. 

\pc{In the datasets, $P$ evaluates to $A$, i.e., is an explanation of $A$.  However, as we show later, if we train a model to predict $A$, and to also predict $P$, the evaluation of $P$ (called $\PAns$) is typically different to $A$. We can view these as two alternative ways to predict an answer, either directly or via explicit program synthesis.
While the synthesis approach is more interpretable, it is not obvious which is better as far as answer prediction is concerned. We evaluate this shortly in Section~\ref{sec:exp}.}



\subsection{Challenge Datasets}

\begin{table*}[ht!]
\resizebox{\textwidth}{!}{
\begin{tabular}{l l}\toprule
    Program & Templated Question \\
    \midrule
    Div(\$y.volume, \$x.volume) & How many \$x fit in \$y \\ 
                                & \eg How many Olympic pools fit in Lincoln Memorial Reflecting pool?\\
    Mul(\$y.density, \$x.volume) & If \$x were to have the same density as \$y, how much would it weigh? \\ 
                                 & \eg If tennis balls were to have the same density as bones, how much would it weigh? \\
    Div(Div(\$y.area, 2), \$x.area) & Assume \$x's area is half its value. How many \$y have the same area as \$x? \\
                               & \eg Assume Indianapolis's area is half its value.  \\
                               & How many Dublin International Airport (DUB) have the same area as Indianapolis?\\
    \bottomrule
\end{tabular}}
\caption{\label{tab:templates} Example templates used for creating the \synthfp dataset along with sample questions for each.}
\vspace{-3mm}
\end{table*}

\subsubsection{\realfp: Real-World Fermi Problems}

The \realfp dataset contains 928 FPs, collected from various internet pages\footnote{including \url{https://www.reddit.com/r/estimation/}}, quizzes, and Fermi problem Olympiads.
The questions cover a wide variety of topics requiring  domain-specific reasoning (such as physics, basic mechanics of Poker, etc), commonsense reasoning,
and most importantly, estimating various physical quantities such as volume, speed, density, etc. 

As discussed in Section~\ref{subsec:dataset-elements}, each instance in \realfp consists of four elements: a question $Q$, an answer $A$ in SI units, supporting facts $F$, and an explanation $P$ in the form of an executable program referring to facts in $F$; \cref{fig:real_example1} shows a sample question from \realfp. 
While $Q$ and $A$ were collected from various sources, $F$ and $P$ were added as part of this work using expert annotation.
It should be noted that the supporting facts and numerical estimates provided in this dataset are a function of the annotator's life experiences and information available on the Internet. As a result, they are not always fully accurate. Due to this, as well as the inherent variance in the answers to FPs,
 our annotations are best viewed as informing us of \emph{one potential way} of approaching the solution.

We split the \realfp dataset into train, validation and test splits containing $185$, $185$ and $558$ questions respectively. 
Reserving a majority (${\sim}60\%$) of FPs for testing is in line with our objective of using the dataset primarily as a test bench to evaluate and drive progress in AI reasoning.
The baseline models we provide use the train set to finetune large-scale models and report performance on the test set. 

\noindent \textbf{Data Analysis.} 
The questions $Q$ in \realfp have a median length of 14 tokens. 
The entire dataset has 892 unique nouns with each question containing 3.7 nouns on average.
Further, the facts and subquestions collected as part of the dataset, contain, on average, an additional ${\sim}4$ nouns.
This indicates that the decomposition for FPs is not trivial and requires recalling or finding information about objects often not mentioned in the original question.
The executable program $P$ provided in the dataset typically contains 2 subquestions; however 176 questions in \realfp~contain a deeper chain of reasoning requiring up to 10 subquestions.

\begin{figure}[t!]
    \begin{center}
\includegraphics[width=1.0\linewidth]{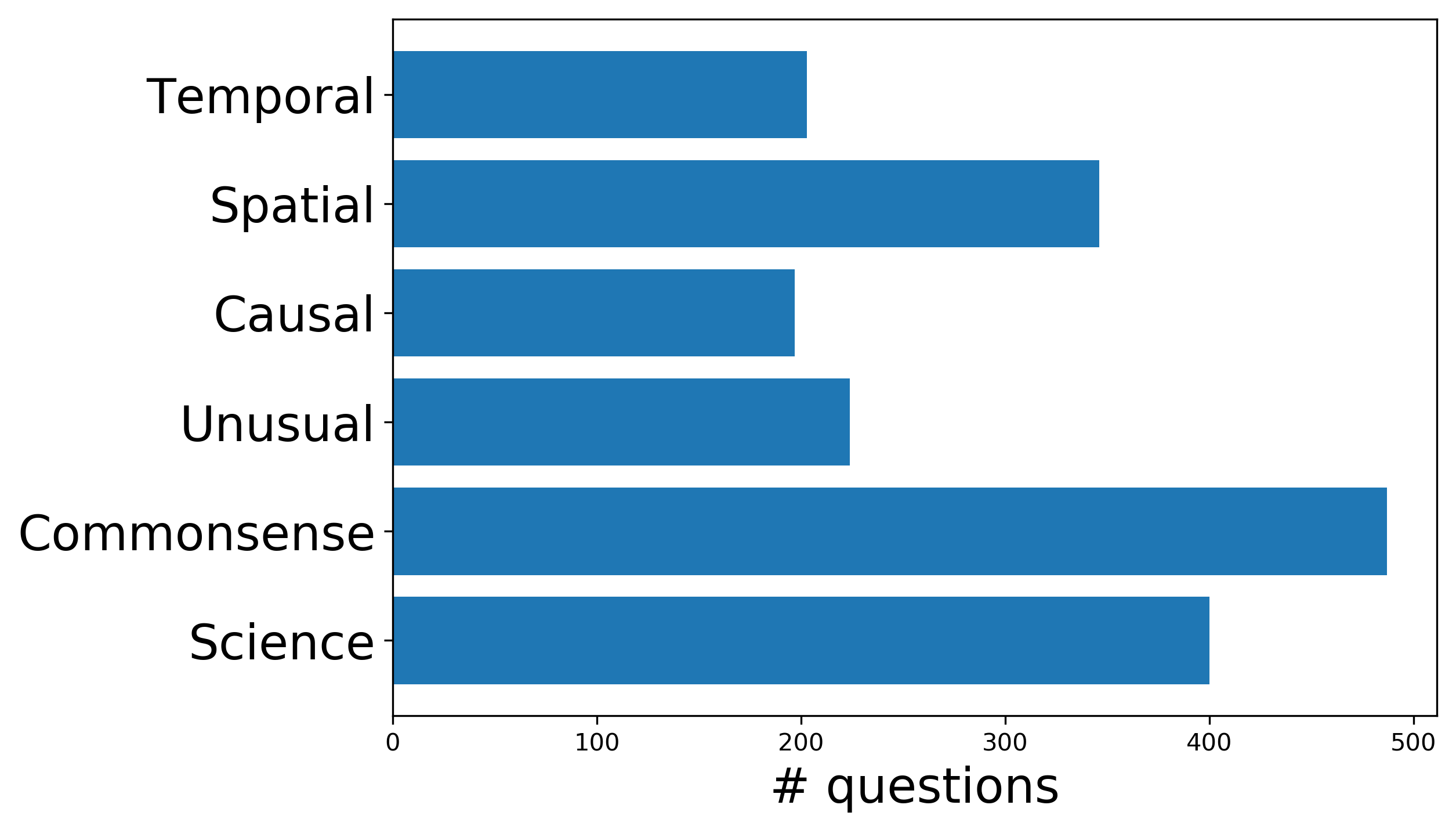}
    \end{center}
\caption{
    Distribution of questions in \realfp based on the type of the reasoning required to arrive at the correct explanation program for a fermi question.
    }
\label{fig:analysis}
\end{figure}

\ak{Further, we analyse the questions in \realfp based on the core reasoning skill required to solve it. 
    For example, the fermi question in \cref{fig:teaser} -- ``How much would the sea level rise if all ice melted?'' is an illustrative example requiring causal and spatial reasoning along with knowledge of science.
    We considered six reasoning types---\emph{spatial} abstraction, \emph{causal} abstraction, \emph{temporal} abstraction, presence of \emph{unusual} attributes or relationships, \emph{commonsense} reasoning and \emph{science}.}
    The frequency of their occurrence in \realfp is summarized in \cref{fig:analysis}.
\ak{Perhaps not surprisingly, commonsense reasoning and science knowledge are required to solve nearly half of the questions. 
    Other reasoning types like abstraction or presence of unusual attributes appear in nearly 25\% of the dataset with potential overlap, \ie one questioning requiring multiple types of reasoning.
}

\subsubsection{\synthfp: Synthetic Fermi Problems}

The complexity of \realfp questions and the relatively small size of the dataset makes it difficult to get started with Fermi-style questions. 
To address this, we introduce a larger dataset of 10k synthetic questions that span a limited set of entities and lines of reasoning, to serve as a sandbox for researchers to help tackle the real-world challenge set. 

After inspecting questions in the \realfp~dataset, we manually selected a few recurring themes to create 12 templates, a few examples of which are shown in \cref{tab:templates}. 
Each template consists of a Fermi-style question with objects represented as variables (\$x, \$y), etc.), and an associated mathematical formula referencing properties of these object variables (e.g., Div(\$y.volume, \$x.volume)).

To illustrate the process of generating a synthetic question from such a template, consider the following FP: ``How many basketballs fit in a schoolbus?". 
The broad template for this question is, ``How many \$x fit in \$y?".
Multiple questions that adhere to this template can be generated by replacing \$x and \$y with objects as long as \$x.volume and \$y.volume are available.
This question generation approach, in addition to ensuring solvability, also provides an easy way to generate an executable program respecting the grammar discussed earlier (e.g., with statements such as Div(Q1, Q2), Q1: ``Volume of Y?'', Q2: ``Volume of X?'', etc.)). 
We provide the full list of 12 templates used to generate the \synthfp dataset in \cref{app:templates}.

Further, we employ the recursive nature of FPs (see \cref{sec:fermi}) to generate more complex solutions for the templated questions in \synthfp. 
For instance, see the last template in \cref{tab:templates}. 
First, the question requires halving the area of Indianapolis and further, in our database, its area is provided in terms of the area of Nauru island.
Therefore solving this question requires a further decomposition \ie ``What is the ratio of the area of Indianapolis and that of Nauru?"
In our dataset, we decompose the solution \wrt another object present in the database for roughly half of the 10k generated templated FPs.

At its core, the synthetic dataset uses a knowledge base $K$, collected via API calls to The Measure of Things resource.\footnote{\url{https://www.themeasureofthings.com}} $K$ contains ${\sim}500$ objects. For each object, it contains information (when applicable and available) about eight common attributes: length, area, volume, weight, density, speed, time, and information (data). 
Starting with $K$, we generate a dataset whose questions have an equal representation of all the templates. 
In total, \synthfp contains 10K FPs with 8K for training, and the remaining 2K FPs equally divided for validation and testing.

\subsubsection{Challenge Tasks \label{subsec:tasks}}

\newcommand{\direst}{\texttt{original}}
\newcommand{\regold}{\texttt{perfect-context}}
\newcommand{\redis}{\texttt{distractor-context}}
\newcommand{\reret}{\texttt{full}}

\pc{We introduce three tasks that each build up to the full complexity of FPs -- allowing researchers to make progress in a measurable and principled manner. For each task, we consider two ways of solving it: (a) generating an answer $A$ directly, where any reasoning is implicit in the parameters of the model, or (b) generating an explanation program $P$, which can then be executed to produce an answer $\PAns$. 
Note that $A$ and $\PAns$ are distinct, reflecting different ways of answering, one directly from the model and one via program synthesis and execution, analogous to ``Thinking, fast and slow" \cite{Kahneman2011ThinkingFA}.}

\paragraph{Task 1, \regold: $Q, F \to A \mid P$.\footnote{Here, $A \mid P$ means that the task can either be treated as that of direct estimation to predict $A$ or as that of synthesis to predict $P$ which can then by executed to produce $\PAns$.}} 
\ak{To help make progress, we define an easier FP task where all and only the relevant facts $F$ are also supplied as the input, along with $Q$.
An example of this I/O is shown in Figure~\ref{fig:real_example1}. \pc{We define two alternative outputs, namely predicting $A$ directly or predicting a program $P$ (which is then evaluated to produce an answer $\PAns$).\footnote{These could be solved using two different models, or one model with two different outputs.}}
}
    
\paragraph{Task 2, \redis: $Q, \{F \cup F_d\} \to A \mid P$.} 
\ak{This setting extends Task 2 by adding $F_d$, a set of distractor facts to the input, bringing the total number of facts to $20$.
This requires the model to also identify which facts are actually useful for the solution. $F \cup F_d$ here is akin to the ``context'' in the typical Reading Comprehension setting studied in the QA literature.}
It should be noted that the set of distractor facts $F_d$ are chosen from facts corresponding to similar questions in the dataset;\footnote{We find that adding random distractors does not make the task harder and systems perform as good as the \regold~setting; indicating no difficulty in identifying the correct facts.} similarity is defined using the question embedding as given by a sentence transformer \cite{reimers-2019-sentence-bert}.
        
\paragraph{Task 3, \reret: $Q \to A \mid P$.} 
\pc{When the input is only the question, we are in the original Fermi problem setting. Again, we define two subtasks (a) generate an answer $A$ directly (b) synthesize a program $P$ which is then used to compute its implied answer $\PAns$.}
Note that when the explanation program $P$ needs to be outputted, the model is not presented with any facts $F$ unlike the previous tasks.
Therefore, the model has the freedom to avail information from any other source -- \eg a knowledge base or via information already part of its parameters.
Given the unconstrained nature of this task, there are many possible programs and facts (the gold program and facts in the dataset represent just
{\it one} possible solution), making fully automatic evaluation out of reach. 
\pc{Instead, we indirectly evaluate $P$ by (a) requiring it to be executable (b) scoring its derived answer $\PAns$ wrt. the gold.
Even these are a high bar due to the challenging nature of FPs.}
Human-in-the-loop
evaluation tools, such as GENIE \cite{genie}, could also be used to directly assess $P$ and $F$
when performance on other metrics reaches a non-trivial level.

\subsection{Metrics}

\paragraph{Answer Evaluation:}
FPs do not have precise answers, because of the underlying ambiguity in terminology and context of both
the original FP and sub-questions which may need to be answered.
Therefore, in Fermi Science Olympiads, participants are awarded full points for obtaining an answer in the same order of magnitude 
as a reference gold answer, and $1/3$ points less for each order of magnitude they are off by. 
In line with this evaluation scheme, we use the following continuous scoring metric:
\begin{equation}
    \fpscore = \max \left\{ 0, 1 - \frac{1}{3}\abs{\log_{10}\frac{A'}{A}}\right\}
\end{equation} 
where $A'$ and $A$ are the predicted and the reference gold answers in SI units, respectively.
During evaluation, we convert the output $A'$ of all models to SI units before comparing with $A$ and therefore, the model is free to output units that are most natural for the question.
The score thus ranges from $1$ when producing precisely the gold reference answer, to $0$ when the prediction is off by three or more orders of magnitude.

\paragraph{Program Evaluation:} 
When we operate in the synthesis setting \ie care about outputting an explanation program $P$ that executes to a numerical estimate $\PAns$, we evaluate explanations (programs) along three axes:
\underline{Validity}: Is the program syntactically valid? This is assessed by seeing whether it successfully evaluates to a number. For this, we use a program executor, written in python,
that evaluates FP programs as described earlier and returns a numerical result, or throws an error.
If execution is successful (independent of the result), validity=1, else 0.

\begin{table*}[t]
\centering
 {\small 
 \setlength{\tabcolsep}{5pt}	
  \begin{tabular}{l l | c | ccc | c | ccc | c | cc} \toprule
    \multirow{3}{*}{\bf Dataset} & \multirow{3}{*}{\bf Model} & \multicolumn{4}{c|}{\bf Task 1: \regold} & \multicolumn{4}{c|}{\bf Task 2: \redis} & \multicolumn{3}{c}{\bf Task 3: \reret} \\ \cmidrule{3-13}
    & & Ans & \multicolumn{3}{c|}{Program $P$}     & Ans & \multicolumn{3}{c|}{Program $P$}     & Ans & \multicolumn{2}{c}{Program $P$} \\
    & & \multicolumn{1}{c}{$A$} & Valid? & $\PAns$ & Facts & \multicolumn{1}{c}{$A$} & Valid? & $\PAns$ & Facts & \multicolumn{1}{c}{$A$} & Valid? & $\PAns$  \\ \midrule
    \synthfp & T5 (FT synth) & 1.00 & 1.00 & 1.00 & 1.00 &0.56 & 0.97 & 0.87 & 0.84 & 0.84 & 1.0 & 0.18 \\ \midrule
\multirow{3}{*}{\realfp} & T5 (FT synth) & 0.14 & 0.27 & 0.14 & 0.85 & 0.06 & 0.33 & 0.04 & 0.6 & 0.13 & \textbf{0.88} & 0.01 \\
& T5 (FT real)  & \textbf{0.18} & 0.71 & \textbf{0.36} & 0.95 & \textbf{0.17} & \textbf{0.89} & 0.21 & 0.75 & \textbf{0.21} & 0.83 & 0.03 \\
& T5 (FT both)  & 0.16 & \textbf{0.76} & \textbf{0.36} & \textbf{0.98} & 0.16 & 0.85 & \textbf{0.23} & \textbf{0.82} & \textbf{0.21} & 0.76 & \textbf{0.04} \\ 
\bottomrule
  \end{tabular}
  }
\caption{Results on FPs with explanations (programs), for T5 fine-tuned on the synthetic FPs (train), the real FPs (train), or
  both. Ans $A$ is the model's direct answer. Explanation (program) $P$ is evaluated
  on whether it executes (Valid?), and if so, whether that execution produces a correct answer ($\PAns$) and whether it uses the
  needed (gold) facts $F$ included in the input for Tasks 1 and 2 (measured as F1 score).}
  \end{table*}
  
\underline{Evaluated Answer accuracy}:
If the program successfully evaluates, how accurate is the resulting evaluated answer?
We use the same answer evaluation metric described earlier. Note that the evaluated program's answer may
(and likely will) differ from the model's direct answer $A$. 
However, if the program is not valid, the model gets a credit of 0.
\ak{Further, it is important to note that this metric assigns credit to outputs that do not necessarily correspond to the explanation program present in the collected dataset. 
As FPs can be solved via multiple possible explanation programs, if a model arrives at the correct answer by such an alternative approach, the evaluated answer accuracy still provides a noisy estimate of the effectiveness of the outputted program.}

\underline{Fact Identification}:
For tasks that include the gold facts $F$ as input (possibly with
distractor facts), did the program $P$ include all and only the gold facts $F$? We compute an F1
measure by comparing the gold fact IDs with the fact IDs used.


%% file: sections/experiments.tex
\section{Experiments}
\label{sec:exp}

We describe some baseline approaches to solve the FP Challenge tasks, and report their performance
on the test sets of the two datasets.

\begin{table*}[t]
\resizebox{\textwidth}{!}{
\begin{tabular}{l c l}\toprule
    $Q$: \textbf{How many people are airborne over Europe at any one moment?} & & \textcolor{blue}{Compiled Answer Accuracy: 0.98} \\ 
    \cmidrule(lr){1-1} \cmidrule(lr){3-3}
    \textcolor{codegreen}{Target Answer: 85090} & & \textcolor{red}{Compiled Answer: 100000} \\ 
    \textcolor{codegreen}{Target Program:} & & \textcolor{red}{Predicted Program:} \\ 
    F1: At any given time, 1270000 people are airborne. & & F1: Around 10000 people are airborne at any moment. \\
    F2: The fraction of Earth as Europe is 6.7\% & & F2: There are around 10 stations in Europe. \\
    Q1: How many people are airborne at any given time? & & Q1: How many flights are there in the US? 
 \\
    Q2: What fraction of Earth does Europe constitute? & & Q2: How many people are airborne at any moment?
\\
    A1: 1270000 & & A1: 10000
\\
    A2: 0.067 & & A2: 10\\
    Q2 -> A2 because  F2, Q1 -> A1 because F1, P: Mul (Q1, Q2) & & Q2 -> A2 because F2, Q1 -> A1 because F1, P: Mul (Q1, Q2)
 \\ 
 \midrule 
 $Q$: \textbf{How many jelly beans fill a bucket?} & & \textcolor{blue}{Compiled Answer Accuracy: 0.00} \\
    \cmidrule(lr){1-1} \cmidrule(lr){3-3}
 \textcolor{codegreen}{Target Answer: 5583.33} & & \textcolor{red}{Compiled Answer: 10}  \\ 
 \textcolor{codegreen}{Target Program: } & & \textcolor{red}{Predicted Program:}  \\ 
 F1: The average volume of a bucket is 0.67 cubic feet. & & F1: The average volume of a bucket is 0.005 cubic feet. \\ 
 F2: The average volume of a jelly-bean is 0.00012 cubic feet. & & F2: The average volume of a jelly-bean is 0.0005 cubic \\
 Q1: What is the average volume of a bucket? & & Q1: What is the average volume of a bucket? \\  
 Q2: What is the average volume of a jelly bean? & & Q2: What is the average volume of a jelly bean? \\ 
A1: 0.67 ft**3 & & A1: 0.005 ft**3\\  
A2: 0.00012 ft**3 & & A2: 0.0005 ft**3 \\
Q2 -> A2 because F2, Q1 -> A1 because F1, P: Div(Q1, Q2) & & Q2 -> A2 because F2, Q1 -> A1 because F1, P: Div(Q1, Q2) \\
\midrule 
 $Q$: \textbf{How many cigarettes have been smoked throughout history?} & & \textcolor{blue}{Compiled Answer Accuracy: 0.00}\\ 
    \cmidrule(lr){1-1} \cmidrule(lr){3-3}
 \textcolor{codegreen}{Target Answer: $1.08\times 10^{15}$} & & \textcolor{red}{Compiled Answer: 100}  \\ 
    \textcolor{codegreen}{Target Program:} & & \textcolor{red}{Predicted Program:} \\ 
    F1: 6.9e+12 cigarettes are sold each year in the world. & &    F1: Around 10 cigarettes have been smoked in the US each year. \\
    F2: It is 157 years since cigarettes were first commercialised. & & F2: An average smoker has been smoked for around 10 years. \\
    Q1: How many cigarettes are sold each year? & & Q1: How many cigarettes have been smoked in the US each year? \\
    Q2: How many years since cigarettes were first commercialised? & & Q2: How many cigarettes have been smoked in the US each year? \\
A1: 6.9e+12 & & A1: 10 \\ 
A2: 157 & & A2: 10 \\
    Q2 -> A2 because F2, Q1 -> A1 because F1, P: Mul (Q1, Q2) & & Q2 -> A2 because F2,  Q1 -> A1 because F1, P: Mul (Q1, Q2) \\ 
 \bottomrule
\end{tabular}}
\caption{\label{tab:qual} Some interesting failure modes of the T5 model (FT both) trained on the \reret~task. 
In the first example presented here, the compiled answer is very close to the target answer as evidenced by the high answer accuracy. However, the decomposition is meaningless. 
On the contrary, the second example is decomposed correctly but the value estimates are significantly different from the true estimate and, therefore, the predicted answer is more than three orders of magnitude farther from the gold answer. 
Finally, we have an example where the decomposition is (trivially) incorrect and, further, the estimates are also significantly different from gold estimates.
}
\end{table*}

\eat{
\begin{table*}[ht!]
\centering
\caption{Results on the \realfp~test set for different FP challenge settings.}\label{tab:results}
\begin{tabular}[t]{lccc}
\hline
& Final Estimate & Compiled Estimate & Support Accuracy \\
\hline
\multicolumn{4}{c}{\regold}\\
\hline
finetuned on \synthfp & 1 & 1 & 1 \\
finetuned on \realfp & 1 & 1 & 1 \\
finetuned on both & 1 & 1 & 1 \\
\hline
\multicolumn{4}{c}{\redis}\\
\hline
finetuned on \synthfp & 1 & 1 & 1 \\
finetuned on \realfp & 1 & 1 & 1 \\
finetuned on both & 1 & 1 & 1 \\
\hline
\multicolumn{4}{c}{\reret}\\
\hline
finetuned on \synthfp & 1 & 1 & n/a \\
finetuned on \realfp & 1 & 1 & n/a \\
finetuned on both & 1 & 1 & n/a \\
\hline
\end{tabular}
\end{table*}
}


\subsection{Baselines} 
These FP Challenge tasks require predicting both the final answer $A$, and the reasoning involved $P$. 
We fine-tune a pre-trained T5 model \cite{raffel2019exploring}\footnote{from the Huggingface library.}
as a seq2seq model that, for each FP challenge task, takes in the corresponding inputs ($Q$ and possibly $F$) to produce the corresponding outputs ($A$, $P$, and possibly $F$).
For each of these tasks, we evaluate the performance on \realfp~after (1) finetuning on the \synthfp-train set, (2) finetuning on the \realfp~train set, and (3) finetuning on the \synthfp~train set in addition to the \realfp-train set.
\\ \\
\noindent \textbf{Results.}
Based on predicted answer ($A$), we find that the T5 model finetuned only on \realfp~performs slightly better than other variants. 
However, \ak{in all the settings}, the best score is achieved when the explanation program is \rc{executed} -- highlighting the utility of outputting the chain of reasoning as opposed to directly predicting an estimate. 
Further, when predicting the program, we observe that fine-tuning on the \synthfp~is useful as it improves other metrics associated with outputting an accurate program (\ie validty and fact f1-measure).
\\ \\
Not surprisingly, the \reret~setting of the FP challenge is significantly challenging and the overall performance is poor compared to other settings where relevant facts are provided to the model.
\rc{As this is the hardest and the most ideal setting of the Fermi problem challenge, we note some observations that may potentially help develop a solver. 
    We observe that a majority of the programs (${\sim}80\%$) are syntactically correct indicating that the poor performance is largely due to lack of semantic understanding.
    While some consistency is found \ie the outputted fact is often relevant to the generated sub-question and the corresponding answer, the decompositions themselves are seldom meaningful\footnote{In some extreme cases, the sub-question itself is not meaningful (\eg ``What is the cost of US?'') and the generated fact can contain wrong units (\eg ``The population of the city is 105 s**2'').}
    -- see \cref{tab:qual} for an instance of such a decomposition.
    And in the rare instance where the generation achieves a partial score, it is likely due to a lucky coincidence -- see \cref{tab:qual} for such a generation.
    Finally, we see that the generations are biased towards simple programs that involve a single operation.
    This is not surprising as the dataset itself contains this bias -- however, the inability of the model to generate complex programs required of the remaining $20\%$ indicates a long-tail problem that can prove challenging in building a solver for Fermi problems.
    In summary, the generations in the \reret~setting demonstrate that the state-of-the-art language models even after sufficient fine-tuning are far from generating semantically meaningful and consistent programs and therefore, Fermi problems are a challenging benchmark that can serve as an indicator of progress in AI reasoning.
}

%% file: sections/conclusion.tex
 \begin{table}[h!t]
  \centering
  \small{ \begin{tabular}{ll} \toprule
    \multicolumn{2}{c}{{\bf Task 3: \reret}} \\ \midrule
    {\bf Model} & {\bf Ans} \\ \midrule
Constant  Prediction    & 0.22  \\
Regression (FT synth) & 0.29  \\ 
Regression (FT real) & 0.13 \\
Regression (FT both) & \textbf{0.32} \\ \bottomrule
  \end{tabular}
  }
  \caption{\label{tab:reg} Performance of the MLP-based regression models for the \reret~task of our FP challenge with results reported on the \realfp~test set. 
}
  \end{table}

\section{Discussion}
  
\subsection{Regression baselines on \reret~task.} We try out some trivial baselines \ie constant prediction and regression to model the \reret~setting of the FP challenge. 
We find interesting trends where even such trivial baselines outperform existing large-scale language-models like T5 on the FP challenge.

\noindent \textbf{Constant Prediction.} This is a trivial baseline that predicts a constant value irrespective of the question.
By performing a logarithmic sweep between $10^{-10}-10^{10}$, we find that the constant prediction of $1000$ (for every FP) achieves an average score of 0.22, indicating that this prediction is, on average, two to three orders of magnitude off.

\noindent \textbf{Regression.} 
This baseline uses a 3-layer MLP, which regresses to a number, given an encoding of the question (obtained using a pre-trained BERT model \cite{devlin2018bert}).
We train this model in three settings: (1) on \synthfp, (2) on \realfp, and (3) on both training sets.
From \cref{tab:reg}, we can see that this model performs best when trained on both datasets (achieving a score of 0.32). However, this is only slightly better than predicting a constant and on average is still off by roughly two orders of magnitude from the correct estimate.

\subsection{Limitations of \realfp.}
\ak{Our \realfp dataset includes only one explanation program to a given FP whereas in practice, there can be multiple creative decompositions that lead to the correct answer.
    To encourage models that are capable of capturing this diversity in the output space, it would be interesting to (a) collect alternative solutions similar to say, image captioning datasets where it is the norm to train and evaluate against multiple ground truth candidates and (b) increasing the number of templates in the \synthfp dataset, thereby biasing the model towards exploring multiple solutions by pre-training on a richer synthetic dataset.
}

Further, the work doesn't include other variants of FPs -- \eg binary yes/no questions, comparisons, or FPs involving probability and risk quantification. Finally, note that our real-world dataset, by virtue of how it is collected, has a high US-centric bias, both in terms of cultural context and vocabulary. 

\subsection{Modeling Improvements.} \ak{In terms of modeling, we establish baselines by finetuning existing large-scale language models.
However, it might be interesting to incorporate them as part of a bigger framework that is developed specially to solve FPs -- for instance, a neuro-symbolic system that intelligently seaches the space of FP decompositions by interleaving question decomposition and the estimation (predicting the numerical answer for sub-questions) phases.
Further, both the estimation phase and the decomposition phase can be improved by giving the model with the ability to access a knowledge base that contains various numerical, commonsense or science facts.
}
\section{Conclusion}

\eat{
\textbf{Limitations.}
In this work, we only consider FPs that have a number as the final answer. 
Hower, there can be other formats of FPs -- \eg binary yes/no FPs (\eg Does the empire state building weigh more than a thousand blue whales?) and comparison FPs (\eg Which is longer? Distance from Earth to Moon or all the spaghetti on Earth arranged in a straight line?). 

Further, this work highlights the reasoning skill of abstraction and proposed FPs as an ideal downstream task for evaluating it. 
Developing datasets tailed to learn to make appropriate abstractions is an interesting direction for future research.

Additionally, it is important to mention that the collected dataset has a strong US-centric bias both in terms of cultural context and vocabulary.}

In this work, we propose Fermi Problems (FPs) as a reasoning challenge for AI systems. 
Apart from introducing \emph{abstraction} as a crucial reasoning skill, our work requires the combined application of various reasoning skills including creative decomposition of problems, commonsense reasoning, mathematical reasoning, \etc.
We collect two datasets -- \realfp with ${\sim}$1k real-world questions and \synthfp with 10k templated questions.
Based on these datasets, we propose three concrete tasks of increasing difficulty that encompass the FP challenge.
The baseline models we provide, despite being based on state-of-the-art language models and even with substantial fine-tuning, struggle on our challenge tasks. They are, on average, off by two orders of magnitude from the correct estimate and perform only slightly better than predicting a constant number.
We thus hope to establish Fermi problems as a hard reasoning challenge that motivates further advances in AI reasoning systems.

%% file: sections/appendix.tex
\onecolumn
\section*{Appendix}
\renewcommand{\thesubsection}{\Alph{subsection}}

\subsection{Templates for \synthfp}
\label{app:templates}

\begin{table*}[h!]
\resizebox{\textwidth}{!}{
\begin{tabular}{l l}\toprule
    Program & Templated Question\\
    \midrule 
    Div(\$y.volume, \$x.volume) & How many \$x fit in \$y? \\
    Div(\$y.length, \$x.length) & How many \$x have the same length as \$y? \\
    Div(\$y.area, \$x.area) & 'How many \$x fit on \$y \\
    Mul(\$k, Div(\$x.data, \$y.data) & How many \$y put together contain the same information as \$k of \$x?\\
    Div(\$y.length, \$x.speed) & How long does it take for \$x to travel across \$y? \\
    Div(\$y.volume, Div(\$x.volume, 2)) & Assume \$y\'s volume is half its value. How many \$x fit in \$y? \\
    Div(\$y.length, Div(\$x.length, 2)) & Assume \$y\'s length is half its value. How many \$x have the same length as \$y?\\
    Div(\$y.area, Div(\$x.area, 2)) & Assume \$y\'s area is half its value. How many \$x fit on \$y?\\
    Div(\$k, \$x.mass) & How many \$x make up \$k kgs?\\
    Mul(\$y.cost, Div(\$k, \$x.cost)) & How many \$x can \$k of \$y buy?\\
    Mul(\$k, Div(\$x.calories / 65)) & How long to digest \$k grams of \$x?\\
    Mul(\$k, Mul(\$y.density, \$x.volume)) & If \$k of \$x were to have the same density as \$y, how much would it weigh\\
    \bottomrule
\end{tabular}}
\caption{\label{tab:all_templates} The templates used to construct \synthfp dataset.}
\end{table*}